\documentclass[conference]{IEEEtran}
\usepackage{amsmath,graphicx}
\usepackage{enumerate}
\usepackage{graphicx}
\usepackage{amsmath,epsfig,amssymb,mathrsfs}
\usepackage{algorithmic}
\usepackage{cite}
\usepackage{amsthm}
\usepackage{arydshln}
\usepackage{amsopn}
\usepackage{color}
\usepackage{verbatim}
\usepackage{amsopn}
\usepackage{amssymb}
\usepackage{algorithm}
\usepackage{enumerate}
\usepackage{subcaption}
\usepackage{setspace} 
\usepackage{float}
\usepackage{multirow}
\usepackage{float}
\usepackage{afterpage}
\usepackage{booktabs}
\usepackage{balance}

\newcommand{\bx}{\mathbf{x}}
\newcommand{\by}{\mathbf{y}}
\newcommand{\bR}{\mathbb{R}}

\newcommand{\bPsi}{\boldsymbol \Psi}
\newcommand{\bPhi}{\boldsymbol \Phi}

\IEEEoverridecommandlockouts

\begin{document}
\title{DeepCodec: Adaptive Sensing and Recovery \\ via Deep Convolutional Neural Networks}
\author{
\IEEEauthorblockN{Ali Mousavi, Gautam Dasarathy, Richard G. Baraniuk  }
\IEEEauthorblockA{
\\Department of Electrical and Computer Engineering \\
Rice University\\
Houston, TX 77005\\
}
\thanks{ This work was supported by NSF CCF-0926127, CCF-1117939; DARPA/ONR N66001-11-C-4092 and N66001-11-1-4090; ONR N00014-10-1-0989, and N00014-11-1-0714; ARO MURI W911NF-09-1-0383.

Email: \{ali.mousavi, gautamd, richb\} @rice.edu }
}

\maketitle

\begin{abstract}
In this paper we develop a novel computational sensing framework for sensing and recovering structured signals. When trained on a set of representative signals, our framework learns to take undersampled measurements and recover signals from them using a deep convolutional neural network. In other words, it learns a transformation from the original signals to a near-optimal number of undersampled measurements and the inverse transformation from measurements to signals. This is in contrast to traditional compressive sensing (CS) systems that use random linear measurements and convex optimization or iterative algorithms for signal recovery. We compare our new framework with $\ell_1$-minimization from the phase transition point of view and demonstrate that it outperforms $\ell_1$-minimization in the regions of phase transition plot where $\ell_1$-minimization cannot recover the exact solution. In addition, we experimentally demonstrate how learning measurements enhances the overall recovery performance, speeds up training of recovery framework, and leads to having fewer parameters to learn.

\end{abstract}

\IEEEpeerreviewmaketitle

\begin{figure*}[h!]
\begin{center}
\includegraphics[width= 15cm]{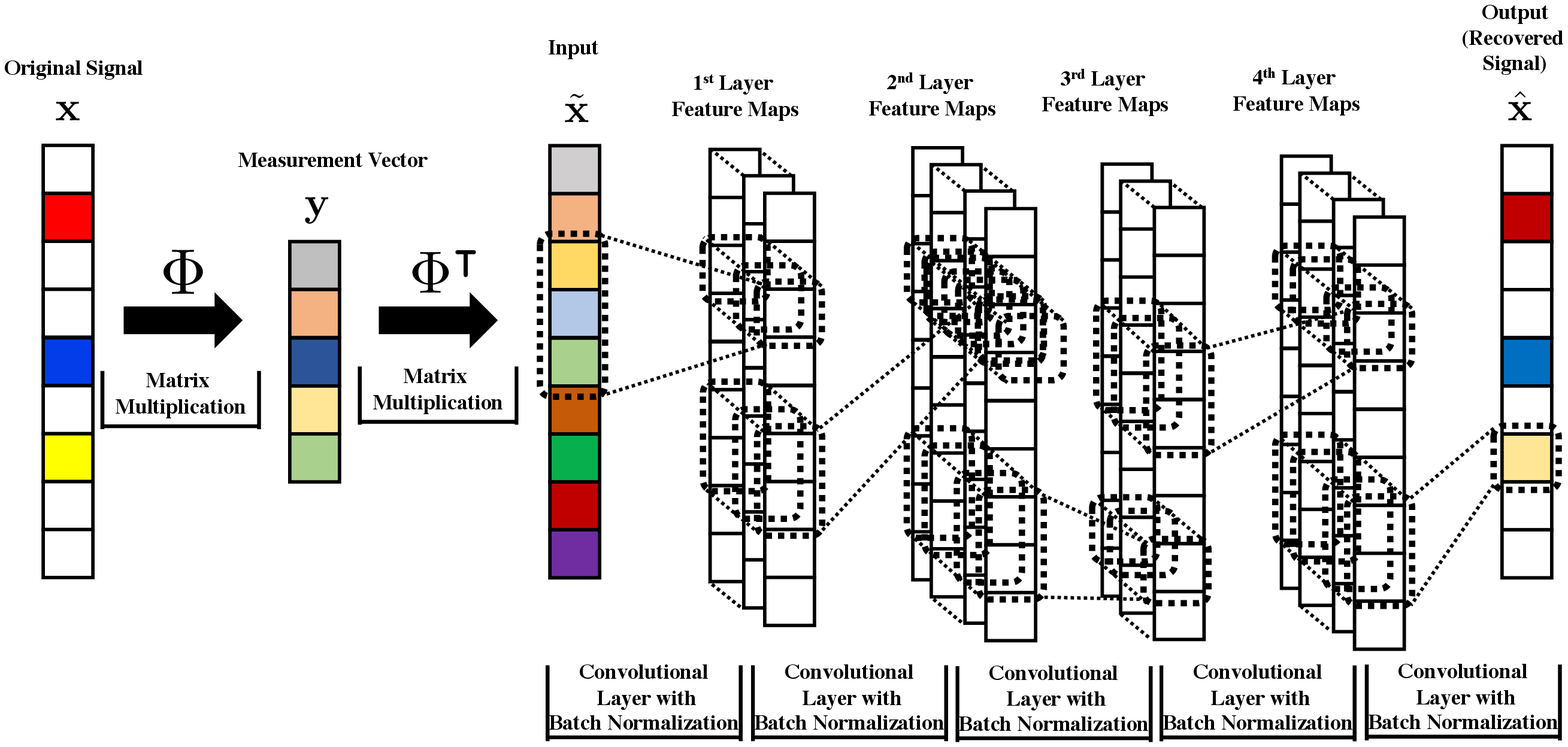}
\caption{
{\em DeepInverse} learns an approximate inverse transformation from measurement vectors $\mathbf{y}$ to signals $\mathbf{x}$ using a deep convolutional
network.
}
\label{fig:DI}
\end{center}
\end{figure*}

\section{Introduction}

Signal recovery is a fundamental problem that appears in a number of important applications. Whenever a signal has some type of structure or redundancy, the signal recovery problem can be solved with a small number of measurements, as compressing the signal appropriately does not lose any information. A widely applicable and well-studied form of signal structure is sparsity. In this case, signal recovery is the problem of recovering a sparse signal $\bx \in \bR^N$ from a set of random linear measurements $\by = \bPhi \bx \in \bR^M$, where $\bPhi$ is an $M\times N$ measurement matrix, and $M$ is typically smaller than $N$.
By sparse, we mean that we can write $\bx = \bPsi \mathbf{s}$, where $\bPsi$ is a basis and only $K \ll N$ of the coefficients $\mathbf{s}$ are nonzero. A natural estimate of the original $\bx$ may be obtained by solving the $\ell_0$ minimization problem, i.e., $\min \|\widehat{\bx}\|_0, ~ \text{s.t.} ~ \by = \bPhi \widehat{\bx}$. However, since solving this optimization program is believed to have combinatorial complexity, several approaches have been proposed to ``relax" it. For instance, arguably the most successful approach is solving the tightest convex relaxation of the above optimization program, i.e., $\ell_1$-norm minimization $ \min \|\widehat{\bx}\|_1, ~ \text{s.t.} ~ \by = \bPhi \widehat{\bx}$ \cite{ChDoSa98,CaTa05}.

The price we pay for using $\ell_1$-minimization instead of $\ell_0$-minimization is reduced recovery performance, namely that $\ell_1$-minimization requires more measurements $M$ to recover a $K$-sparse signal than $\ell_0$-minimization.
Let $\delta = \frac{M}{N}$ denote the undersampling ratio and let $\rho = \frac{K}{M}$ indicate the normalized sparsity level. 
An instructive way of studying these phenomena is to visualize the probability of successful recovery for each value of $(\delta,\rho) \in [0,1]^2$. Given that there is a sharp \emph{phase transition} in the recovery performance of $\ell_1$-minimization, such plots have come to be known as phase transition plots \cite{DoTa05} (see Figure \ref{fig:PT} for an example of such a plot). The two-dimensional {\em phase transition} plot has two phases: a success phase and a failure phase, where $\ell_1$-minimization can and cannot recover the exact signal, respectively. In other words, $\ell_1$-minimization successfully recovers the sparse signal if its normalized sparsity level is less than a certain threshold. 

Our goal in this paper is to show that by using deep learning techniques, we can design computational sensing frameworks that can overcome the limitations of $\ell_1$-minimization. To achieve this goal, we make improvements to $\ell_1$-minimization in 2 separate directions:

\begin{itemize}
\item First, instead of traditional schemes that use random undersampled measurements we \emph{learn} a transformation from original signals to undersampled measurements. The critical issue with random undersampled measurements is that they are universal and do not use the structure that is specific to the set of signals that we care about for the particular problem at hand. However, one can use these structures to find a better way to compress signals and derive measurements. 
\item Second, instead of using iterative or convex optimization algorithms, we learn the inverse transformation from undersampled measurements to original signals using a deep convolutional network (DCN) and training it on several examples. 
\end{itemize}
Our new framework is thus able to learn an efficient and compressed representation of training signals, and also the corresponding inverse map from this representation onto the original signal space. We call our framework {\em DeepCodec}, where codec of course is the well known portmanteau of  ``coder-decoder''.  {\em DeepCodec} is closely related to the {\em DeepInverse} framework \cite{deepinverse} proposed by a subset of the current authors. 
The major difference between {\em DeepInverse} and {\em DeepCodec} is that {\em DeepInverse} recovers original signals from random linear undersampled measurements while {\em DeepCodec} learns to take nonlinear undersampled measurements and recovers original signals from them.  

In this paper, we study the sparse recovery performance of {\em DeepCodec} and show that it significantly outperforms $\ell_1$-minimization in this context. In particular, {\em DeepCodec} succeeds, with overwhelmingly high probability, even when the normalized sparsity of the problem at hand is larger than the threshold that comes from $\ell_1$-minimization's phase transition (i.e., in regimes where $\ell_1$-minimization almost always fails). We show that {\em DeepInverse} also has this feature. We also compare these algorithms in terms of their runtime
and show that {\em DeepCodec} and {\em DeepInverse} are orders of magnitude faster than the conventional algorithms. The tradeoff for the ultrafast runtime is a one-time, computationally intensive, off-line training procedure typical of deep networks. 
This makes our approach applicable to real-time sparse recovery problems. Finally, we show how taking adaptive measurements and learning a transformation from signals to their undersampled measurements improves recovery performance compared to simply using random measurements. In other words, we show adaptivity helps {\em DeepCodec} to outperform {\em DeepInverse}.

The rest of this paper is organized as follows: Section \ref{sec:priorWork} summarizes prior  art in using deep learning frameworks for structured signal recovery. Section \ref{sec:DeepCodec} introduces the network architecture we have used to take adaptive measurements from signals and to recover signals using adaptive measurements. Section \ref{sec:simul} summarizes our experimental results and the comparison of our method with previous works. We make some concluding remarks in Section~\ref{sec:con}. 

\begin{figure*}[h!]
\begin{center}
\includegraphics[width= 14cm]{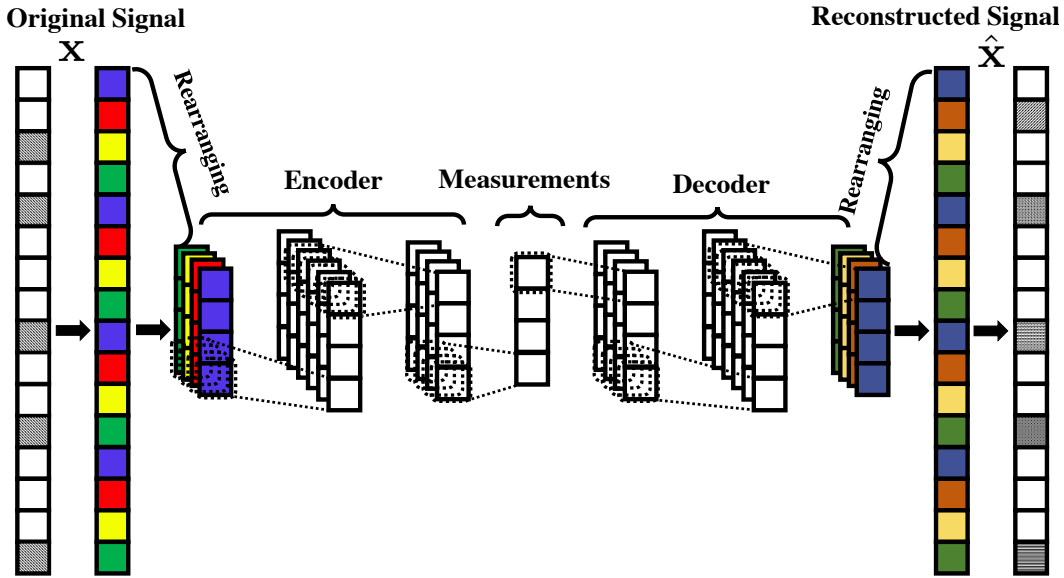}
\caption{
{\em DeepCodec} learns a transformation from signals $\mathbf{x}$ to measurement vectors $\mathbf{y}$ and an approximate inverse transformation from measurement vectors $\mathbf{y}$ to signals $\mathbf{x}$ using a deep convolutional
network that consists of convolutional and sub-pixel convolution layers.
}
\label{fig:CAE}
\end{center}
\end{figure*}

\section{Prior Work} \label{sec:priorWork}
In this section we briefly describe previous works on designing structured signal recovery algorithms. These algorithms span a wide spectrum but may be itemized as the follows:
\begin{itemize}
\item \textit{Physics-driven approaches}. These methods are conventional recovery algorithms inspired by the concept of sparsity. They are mainly based on enforcing sparsity (or other types of structure) in the signal approximation/estimation process. The most well-known methods from these approaches are greedy algorithms \cite{TroppNeedellCoSaMP}, convex optimization algorithms \cite{CaTa05}, and iterative algorithms \cite{DoMaMo09,ISTA}.
\item \textit{Mixtures of physics-driven and data-driven approaches}. This class of algorithms mixes concepts from physics-driven approaches that give them interpretability and the power of using training data that gives them adaptivity and performance enhancement for specific applications. In most cases, these approaches are a combination of conventional sparse recovery algorithms and deep learning frameworks \cite{metzler2017learned,LAMP,kamilov2016learning,OneNettoSolve,LISTA}
\item \textit{Data-driven approaches}. These methods are mainly based on designing frameworks that can use training data to learn a representation of signals and a transformation from undersampled measurements to original signals \cite{mousaviSDA,reconnet,deepinverse,dr2net}. The main benefit of using these approaches is that they provide ultrafast signal recovery. However, this ultrafast runtime comes at a price that is a computationally intensive, off-line training procedure typical to deep networks. Since this training procedure needs to be completed only once, data-driven approaches are applicable for real-time signal recovery problems.

\end{itemize}

The first data-driven signal recovery approach that was based on deep learning frameworks was introduced in \cite{mousaviSDA}. Authors in \cite{mousaviSDA} used stacked denoising autoencoders (SDA) as an unsupervised feature learner. The major drawback of using SDA is that it consists of fully-connected layer. This means that as signal size grows, we need to tune a larger set of network parameters and this could cause overfitting if we do not have enough training data. One solution that authors introduced in \cite{mousaviSDA} was dividing signals into smaller blocks and performing blocky reconstruction. This approach is not applicable for certain applications like compressive sensing MRI where measurements correspond to the whole signal and one cannot divide signals into smaller non-overlapping or overlapping blocks.

As an alternative to the SDA approach, a subset of the authors of this paper in \cite{deepinverse} introduced {\em DeepInverse} that is a signal recovery framework based on DCNs. The main motivation for replacing the SDA approach with DCNs is their two distinctive features: first, sparse connectivity of neurons in each layer. Second, having shared weights across the entire receptive fields. These two features significantly reduce computational complexity of DCNs compared to the SDA approach and make DCNs distinctively applicable for structured signal recovery problem. In the following, we briefly describe the {\em DeepInverse} framework \cite{deepinverse} for sparse signal recovery (see Figure \ref{fig:DI}).

{\em DeepInverse} takes as input a set of measurements $\mathbf{y}$ in $\mathbb{R}^M$ and outputs the signal estimate $\widehat{\mathbf{x}}$ in $\mathbb{R}^N$. 
To increase the dimensionality of the input from $\mathbb{R}^M$ to $\mathbb{R}^N$, it applies the adjoint operator $\boldsymbol \Phi^{\intercal}$ in the first layer. 
To preserve the dimensionality of the processing in $\mathbb{R}^N$, it dispenses with the downsampling max-pooling operations made popular in modern DCNs \cite{lecun1998gradient}. 
We assume that the measurement matrix $\bPhi$ is fixed. 
Therefore, each $\mathbf{y}_i$ ($1 \leq i \leq M$) is a linear combination of $\mathbf{x}_j$s ($1\leq j \leq N$). 
By training a DCN, we learn a nonlinear mapping from the signal proxy $\mathbf{\tilde{x}}=\boldsymbol \Phi^{\intercal} \by$ to the original sparse signal $\bx$. 

Among the many possibilities for the deep network architecture,  {\em DeepInverse} uses one layer to implement the adjoint operator $\boldsymbol \Phi^{\intercal}$ and five convolutional layers with their corresponding batch normalization \cite{ioffe2015batch} layers.  
Each convolutional layer applies a leaky-ReLU \cite{maas2013rectifier} nonlinearity to its output. 
The $i$-th entry of the $t$-th feature map in the first convolutional layer receives the signal proxy $\mathbf{\tilde{x}}$ as its input and outputs $(\bx_{c_1})_{i}^t=\mathcal{S}(\text{L-ReLU}((\mathbf{W_1^t} \ast \mathbf{\tilde{x}})_{i}+(\mathbf{b_1^t})_{i}))$, where $\mathbf{W_1^t} \in \mathbb{R}^{P}$ and $\mathbf{b_1^t} \in \mathbb{R}^{N+P-1}$ denote the filter and bias values corresponding to the $t$-th feature map of the first layer and $\text{L-ReLU}(x) = x$ if $x > 0$ and $=0.01x$ if $x \leq 0$. 
Finally, the subsampling operator $\mathcal{S}(\cdot)$ takes the output of $\text{L-ReLU}(\cdot)$ to the original signal size by ignoring the borders created by zero-padding the input. 
The feature maps for the other convolutional layers are processed in a similar manner. 
If we denote the set of weights and biases in the DCN by $\Omega$, then we can define a nonlinear mapping from the measurements to the original signal by $\mathbf{\widehat{x}}=\mathcal{M}(\mathbf{y},\Omega)$. 
To learn the weights and biases, we employ backpropagation algorithm to minimize the mean-squared error (MSE) of the estimate $\mathbf{\widehat{x}}$.

\section{DeepCodec} \label{sec:DeepCodec}
Having reviewed the {\em DeepInverse} framework, we are now ready to introduce our new framework {\em DeepCodec} (See Figure \ref{fig:CAE}). Similar to {\em DeepInverse}, its sequel {\em DeepCodec} learns the inverse transformation from undersampled measurements vectors to signals using a special form of DCNs. However, there is a major difference between {\em DeepInverse} and {\em DeepCodec}. {\em DeepInverse} uses adjoint operator $\boldsymbol \Phi^{\intercal}$ in the first layer to make a proxy of original signals from their random linear undersampled measurements. However, {\em DeepCodec} does not use random linear undersampled measurements of original signals. Instead, it learns to take nonlinear undersampled measurements from original signals and recover original signals from these learned measurements. Learning measurements from original signals helps to preserve more information compared to taking random measurements and we experimentally show this in Section \ref{sec:simul}.

For taking undersampled measurements from original signals, one should reduce their dimensionality. In DCNs, the conventional form of dimensionality reduction is by using pooling layers that are a form of downsampling and provide translation invariance. There are different ways of implementing pooling layers such as max pooling where we take the maximum value from each cluster of neurons; or average pooling where we take the average of values in each cluster. In addition to these traditional implementations of a pooling layer, there are new approaches for implementing it such as stochastic pooling \cite{zeiler2013stochastic} and spatial pyramid pooling \cite{he2014spatial}.

Almost all implementations of a pooling layer are hand-designed in the sense that either they are designed for a special goal like achieving translation invariance or they use some sort of expert knowledge. Since they are hand-designed, they do not optimally preserve signal information while reducing its dimensionality. Our argument here is that one can learn dimensionality reduction from data instead of using a hand-designed transformation. In other words, learning dimensionality reduction from data itself, if done correctly, will preserve more information compared to conventional ways that do not use training data for downsampling. 

In this paper, instead of using conventional implementations of pooling layers, we learn a transformation for dimensionality reduction. The major benefit of learning this transformation is that we preserve more information. The tradeoff for the preservation of information is a computationally more complex pooling layer compared to conventional ones such as max-pooling or average-pooling. The design of our pooling layer is inspired by the sub-pixel convolution layer initially introduced in \cite{shi2016real}. Sub-pixel convolution layer was initially designed for aggregating feature maps from low-resolution image space in order to build a high-resolution image. In other words, it was initially designed for increasing the dimensionality rather than reducing it. However, as we show in the following, with some modifications it can be used for dimensionality reduction as well. Once we reduce signal dimensionality and take undersampled measurements, we can use convolutional layers plus a sub-pixel convolution layer to reconstruct signals from their undersampled measurements. Having said this preliminary explanation of {\em DeepCodec}, we are now ready to describe it in more detail.

{\em DeepCodec} takes an input that is an original signal $\bx \in \bR^N$. For the sake of brevity we assume that {\em DeepCodec} input is a single channel and one-dimensional signal. However, it is straightforward to extend it to multi-channel and multi-dimensional signals like RGB images. In addition, we assume that the length of original signals $N$ is divisible by the length of undersampled measurements vector we are interested in $M$, i.e., $N=rM$. We should note that it is straightforward to extend {\em DeepCodec} to cases where $N$ is not divisible by $M$.  For {\em DeepCodec} the first task is to take undersampled measurements from its input. To do so, the first layer of {\em DeepCodec} rearranges its input. Let the size of input layer be $N\times1\times1$ where $N$ denotes the length, the first 1 denotes the width, and the second 1 denotes the number of input channels. The rearranging layer takes this input and turns it into an $M\times1\times r$ signal where $M=N/r$. In other words, this rearranging layer divides the length of output signal by a factor of $r$ while increase the number of channels in output signal by a factor of $r$. Mathematically, we can describe a rearranging layer as the following:
\begin{align}\label{eq:ds}
\tilde{\bx}(\bx,r)_{i,1,c} = \bx_{ i\times r + mod(c,r),1,1},
\end{align}
where $\tilde{\bx}$ denotes the output of the rearranging layer and $\bx$ denotes its input. This rearranging layer clearly reduces its input dimensionality; however, its output is still not an undersampled measurements vector of the original signal since the total number of output neurons is the same as input layer. Therefore, we employ several convolutional layers to adjust the total number of neurons such that it equals to $M$ that is the number we are interested in. We can mathematically formulate the $i$-th entry of the $t$-th feature map of the $l$-th convolutional layer as the following:
\begin{align}\label{eq:conv}
(\bx_{l})_{i,1,t}=\mathcal{S}(\text{ReLU}(\mathbf{W_l} \ast \bx_{l-1}+\mathbf{b_l}))_{i,1,t},
\end{align}
where $\bx_{l-1} \in \bR^{M\times 1 \times c_{l-1}}$ denotes the input of the $l$-th convolutional layer and $c_{l-1}$ denotes the number of channels in $\bx_{l-1}$. $\mathbf{W_l} \in \bR^{h_{f_l}\times 1 \times c_{l-1} \times c_{l} }$ and $\mathbf{b_l} \in \bR^{c_{l}}$ denote the filter and bias values corresponding to the $l$-th convolutional layer where $h_{f_l}$ is the length of the $l$-th convolutional layer's filter. Finally, $\bx_{l} \in \bR^{M\times 1 \times c_{l}}$ denotes the output of the $l$-th convolutional layer. If the $p$-th layer's output corresponds to undersampled measurements, then $\mathbf{W_p} \in \bR^{h_{f_p}\times 1 \times c_{p-1} \times 1 }$ and  $\bx_p \in \bR^{M\times 1 \times 1}$. Finally, $\text{ReLU}(x) = \max(0, x)$ and the subsampling operator $\mathcal{S}(á)$ takes the output of $\text{ReLU}(.)$ to the original signal size by ignoring the borders created by zero-padding the input.

Once we have the undersampled measurements vector (i.e., the output of $p$-th layer), we employ several convolutional layers to extract feature maps from it. The main advantage of this method compared to {\em DeepInverse} is that in {\em DeepInverse} we were reconstructing signals from their proxies which had the same size as original signals. However, in  {\em DeepCodec} we are reconstructing signals in measurement domain by employing several convolutional layers that receive undersampled measurements as their initial input. 

The output of the $p$-th layer (i.e., undersampled measurements vector) lies in $\bR^{M\times 1 \times 1}$. However, the reconstructed signal should lie in $\bR^{N\times 1 \times 1}$ where $N=r\times M$. Therefore, we have to boost the dimensionality from $\bR^{M\times 1 \times 1}$ to $\bR^{(r\times M)\times 1 \times 1}$. We initially employ several convolutional layers as we introduced in \eqref{eq:conv}. These convolutional layers will help us to boost the dimensionality from $\bR^{M\times 1 \times 1}$. In particular since the output should lie in $\bR^{(r\times M)\times 1 \times 1}$, we boost the dimensionality of the $p$-th layer from $\bR^{M\times 1 \times 1}$ to $\bR^{M\times 1 \times r}$ through employing several convolutional layers in measurements domain. Once we have boosted the dimensionality of the measurements vector to $\bR^{M\times 1 \times r}$, we employ a sub-pixel convolution layer \cite{shi2016real} to rearrange neurons and produce an output which lies in $\bR^{N\times 1 \times 1}$ from its input which lies in $\bR^{M\times 1 \times r}$. Mathematically, we can describe our sub-pixel convolution layer as the following:

\begin{align}
\hat{\bx}(\bx,r)_{i,1,1} = \bx_{ (i/r),1,mod(i,r)},
\end{align}
where $\hat{\bx}$ denotes the output of the rearranging layer and $\bx$ denotes its input.

Figure \ref{fig:CAE} shows the schematic of {\em DeepCodec} framework. Note that we can consider {\em DeepCodec} as a special form of convolutional autoencoder. Here is the summary of how {\em DeepCodec} works:
\begin{itemize}
\item Receiving an input signal.
\item Rearranging input's components and reduce its dimensionality through convolutional layers.
\item Taking undersampled measurements.
\item Boosting measurements dimensionality through convolutional layers.
\item Transforming the output to a reconstructed signal through a sub-pixel convolution layer.
\end{itemize}
If we denote the output of {\em DeepCodec} by $\hat \bx$ and assume that {\em DeepCodec} has overall of $d$ convolutional layers, then we can denote its set of parameters by $\Omega = \{\mathbf{W}_j, \mathbf{b}_j\}_{j=1}^d$ and define a nonlinear mapping from original signals to reconstructed signals as $\hat \bx = \mathcal{F}(\bx, \Omega)$. Now if we have a training set $\mathcal{D}_{\text{train}} = \{\bx^{(i)}\}_{i=1}^{s}$ that consists of $s$ original signals, we can use the mean squared error (MSE) as a loss function over the training set $\mathcal{D}_{\text{train}}$
\begin{align}
\mathcal{L}(\Omega) = \frac{1}{s} \sum_{i=1}^s \|\mathcal{F}(\bx^{(i)}, \Omega)-\bx^{(i)}\|_2^2.
\end{align}
we can employ either stochastic gradient descent (SGD) or ADAM optimizer \cite{ADAMopt} to minimize $\mathcal{L}(\Omega)$ and learn weights and biases.

\section{Experimental Results} \label{sec:simul}

In this section, we compare the performance of {\em DeepCodec} to {\em DeepInverse} and to the LASSO \cite{TibLasso96} $\ell_1$-solver (implemented using the coordinate descent algorithm of \cite{friedman2010regularization}) over a grid of regularization parameters. In all experiments, we assume that the optimal regularization parameter of LASSO is given by an oracle.

Our {\em DeepCodec} framework has eight layers. The first layer is a rearranging layer for dimensionality reduction. We assume that the output of the first layer lies in $\bR^{M\times1\times r}$. The second to seventh layers are all convolutional layers and the size of filters in all of these layers are $49\times1$. The second layer has 8 filters each having $r$ channels of size $49\times1$. The third layer has 4 filters each having 8 channels of size $49\times1$. The fourth layer has 1 filter that has 4 channels of size $49\times1$. The output of the fourth layer is our undersampled measurements vector. The fifth layer has 4 filters each having 1 channel of size $49\times1$. The sixth layer has 8 filters each having 4 channels of size $49\times1$. The seventh layer has $r$ filters each having 8 channels of size $49\times1$. Finally, the eighth layer is a sub-pixel convolution layer for aggregating feature maps of the seventh layer and boosting the dimensionality. The eighth layer gets an input which lies in $\bR^{M\times 1 \times R}$ and converts it into a signal reconstruction that lies in $\bR^{N\times 1 \times 1}$.

Our {\em DeepInverse} network has five layers. 
The first and third layers have 32 filters, each having 1 and 16 channels of size $125\times 1$, respectively. 
The second and fourth layers have 16 filters, each having 32 channels of size $125\times 1$. 
The fifth layer has 1 filter that has 16 channels of size $125\times 1$. 
We trained and tested {\em DeepCodec} and {\em DeepInverse} using wavelet sparsified versions of 1D signals of size $N=512$ extracted from random rows of CIFAR-10 images \cite{krizhevsky2009learning}. 
The training set contains 100,000 signals, and the test set contains 20,000 signals.

\begin{figure}[h!]
\begin{center}
\includegraphics[width= 9cm]{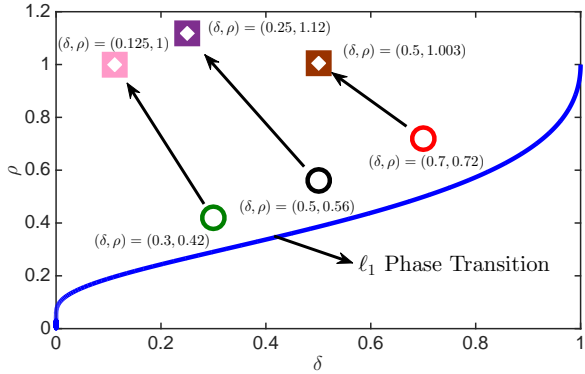}
\caption{The blue curve is the $\ell_1$ sparse recovery phase transition. The circular and square points denote problem configurations for {\em DeepInverse} and {\em DeepCodec} respectively. Arrows between circular and square points show different configurations for the same set of signals. As shown in Table \ref{tab:nmse} {\em DeepCodec} outperforms {\em DeepInverse} in recovering the same signals even when the configuration is more challenging.}
\label{fig:PT}
\end{center}
\end{figure}

The blue curve in Figure \ref{fig:PT} is the $\ell_1$ phase transition curve. The circular points denote the problem instances, i.e., $(\delta,\rho)$, on which we study the performance of {\em DeepInverse} and the LASSO. By design, these problem instances are on the ``failure'' side of the $\ell_1$ phase transition. The undersampling ratios, i.e., $\delta$ for these instances are 0.3, 0.5, and 0.7 and the normalized sparsity, i.e., $\rho$ for these instances are 0.42, 0.56, and 0.72 respectively. 

We have used the same set of sparse signals for training and testing {\em DeepCodec} framework to compare its performance with {\em DeepInverse} and the LASSO. The square points in Figure \ref{fig:PT} denote the problem instances, i.e., $(\delta,\rho)$, on which we have trained and tested {\em DeepCodec}. As we mentioned, signals that we have used for these problem instances are the same ones we have used for training and testing {\em DeepInverse} with settings denoted by circular points. However, for {\em DeepCodec} we have made recovery problems harder by reducing undersampling ratios. The arrows between square points and circular points in Figure \ref{fig:PT} denote correspondence between problem instances in {\em DeepCodec} and {\em DeepInverse}. As an example, the set of $k$-sparse signals in $\bR^N$ where $k=64$ and $N=512$ corresponds to $(\delta,\rho)=(0.3,0.42)$ for {\em DeepInverse} (with undersampling ratio 0.3) while corresponds to $(\delta,\rho)=(0.125,1)$ for {\em DeepCodec} (with undersampling ratio 0.125). 

Table \ref{tab:nmse} shows the average normalized mean squared error (NMSE) for the test set signals using all three methods. As we can see in this table, even though we have made recovery of the same signals significantly harder for {\em DeepCodec} by reducing the undersampling ratio, it still outperforms {\em DeepInverse} and the LASSO (with the optimal regularization parameter) in all of the configurations determined in Figure \ref{fig:PT}.

Table \ref{tab:numParams} compares the number of parameters learned in each setting for {\em DeepInverse} and {\em DeepCodec}. As we can see, in all of the problem instances {\em DeepCodec} has significantly fewer number of parameters compared to {\em DeepInverse} while outperforming it in recovery performance. This is mainly due to the fact that  {\em DeepCodec} learns a transformation for taking measurements from signals while {\em DeepInverse} uses random measurements.

Table \ref{tab:runtime} compares the computational complexity of all three methods. If we use a fast iterative algorithm for solving the LASSO like AMP \cite{mousavi2015consistent}, then the runtime of every iteration would be $\mathcal{O}(MN)$ where $N$ is the size of signal and $M$ is the size of undersampled measurements vector. If we let $M=ck\log(N)$, then this runtime would be $\mathcal{O}(kN\log(N))$.  In {\em DeepInverse} every convolutional layer's input and output has the same size as the input signal that is $N$ and hence, runtime of computing every layer's output is $\mathcal{O}(N)$. Therefore, computational cost corresponding to one layer of {\em DeepInverse} is significantly less than the one for an iteration of AMP for solving LASSO. In {\em DeepCodec} since we are recovering signals in the measurements space, computing the output of typical middle layers will cost $\mathcal{O}(M)$. As we can see, not only {\em DeepCodec} gives us a better recovery performance compared to {\em DeepInverse}, but also it has a faster runtime and needs fewer parameters to learn. It is noteworthy to mention that usually the number of iterations an iterative algorithm such as AMP needs for recovering a signal is tens of times more than the number of convolutional layers needed by either {\em DeepInverse} or {\em DeepCodec} to recover the same signal. This is another factor that makes signal recovery by {\em DeepInverse} and {\em DeepCodec} significantly faster than iterative algorithms like AMP.

\begin{figure}[h!]
\begin{center}
\includegraphics[width= 9cm]{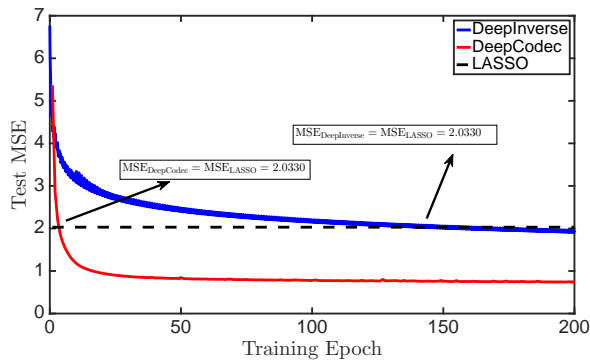}
\caption{Test MSE of {\em DeepInverse} and {\em DeepCodec} during training epochs for $(\delta,\rho) = (0.7,0.72)$. {\em DeepCodec} outperforms LASSO after only 4 epochs while {\em DeepInverse} outperforms LASSO after 138 epochs.}
\label{fig:epoch}
\end{center}
\end{figure}

Figure \ref{fig:epoch} compares the effect of training on {\em DeepInverse} and {\em DeepCodec}. It shows the MSE of recovering test signals by {\em DeepInverse} and {\em DeepCodec} in different training epochs. The set of training signals are the same for both {\em DeepInverse} and {\em DeepCodec}. However, the problem instance for {\em DeepInverse} is $(\delta,\rho)=(0.7,0.72)$ while for {\em DeepCodec} is $(\delta,\rho)=(0.5,1.003)$ which means we have given {\em DeepCodec} a more difficult recovery problem. However, as we can see in Figure \ref{fig:epoch}, training is significantly faster for {\em DeepCodec} compared to {\em DeepInverse}. {\em DeepCodec} outperforms the LASSO (for the problem instance $(\delta,\rho)=(0.7,0.72)$) after only 4 training epochs while for {\em DeepInverse} it takes 138 epochs to outperform the LASSO. This fast training has two major reasons: First, {\em DeepCodec} has fewer number of parameters to learn. Second, {\em DeepCodec} learns adaptive measurements instead of using random measurements.

\begin{figure}[h!]
\begin{center}
\includegraphics[width= 8.3cm]{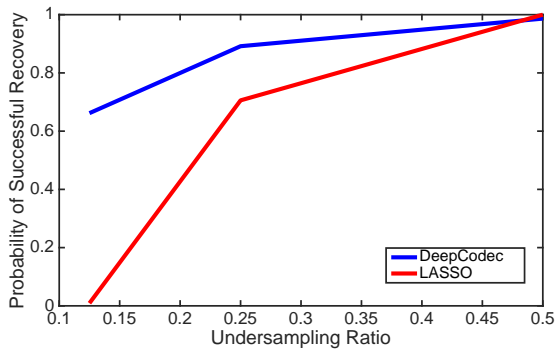}
\caption{ Average probability of successful signal recovery for different undersampling ratios when we use {\em DeepCodec} and LASSO. This plot studies three different configurations: below, above, and on the $\ell_1$ phase transition curve. }
\label{fig:suc}
\end{center}
\end{figure}

Figure \ref{fig:suc} compares the probability of successful recovery by {\em DeepCodec} and LASSO as measured by 20000 Monte Carlo samples (that are test set signals). For each undersampling ratioÊ $\delta$ and for the $j$-th Monte Carlo sample, we define the success variable $\varphi_{\delta,j}=\mathbb{I}\left(\frac{\| \hat{\bx}^{(j)}-\bx^{(j)} \|_2^2}{\| \bx^{(j)} \|_2^2} \le 0.01 \right)$, where $\bx^{(j)}$ is the $j$-th Monte Carlo sample, $\hat{\bx}^{(j)}$ is the recovered signal from measurements of $j$-th Monte Carlo sample, and $\mathbb{I}(.)$ is the indicator function. We define the empirical successful recovery probability as $P_{\delta} = \frac{1}{q} \sum_{j=1}^q \varphi_{\delta,j}$, where $q$ is the total number of Monte Carlo samples. In Figure \ref{fig:suc}, our test set signals are $k$-sparse in $\bR^N$ where $k=34$ and $N=512$ and we have considered three different configurations: 
\begin{itemize}
\item $M=64$ measurements which corresponds to $(\delta,\rho)=(0.125,0.53)$ that is a setting above the $\ell_1$ phase transition, i.e., failure side.
\item $M=128$ measurements which corresponds to $(\delta,\rho)=(0.25,0.26)$ that is a setting on the $\ell_1$ phase transition curve.
\item $M=256$ measurements which corresponds to $(\delta,\rho)=(0.5,0.13)$ that is a setting below the $\ell_1$ phase transition, i.e., success side.
\end{itemize}
As we can see in Figure \ref{fig:suc}, {\em DeepCodec} significantly outperforms LASSO when the problem configuration lies above or on the $\ell_1$ phase transition curve. Only when the problem configuration lies below the $\ell_1$ phase transition curve LASSO slightly outperforms {\em DeepCodec} (probability of successful recovery equals to 1 for LASSO vs. 0.99 for {\em DeepCodec}). This is expected since for a setting below $\ell_1$ phase transition curve, we expect $\ell_1$ minimization to behave the same as $\ell_0$ minimization. In other words, we expect to have a successful recovery based on physics of $\ell_1$ minimization. However, {\em DeepCodec} should learn a transformation for transforming measurements back to the original signals. In this setting, we have trained {\em DeepCodec} for only 5000 epochs. We conjecture that if we train {\em DeepCodec} for more than 5000 epochs or use more number of parameters, it can achieve probability of successful recovery equals to 1. We leave this topic, i.e., understanding the number of parameters and training epochs needed for achieving certain recovery quality, as an avenue for the future research.

\begin{figure}[h!]
\begin{center}
\includegraphics[width= 9.5cm]{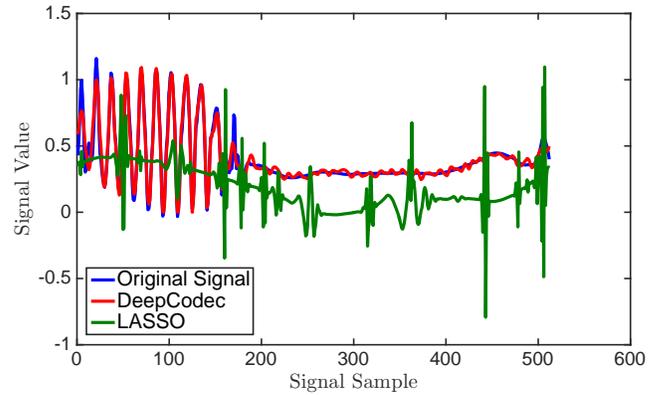}
\caption{Recovery example by {\em DeepCodec} and LASSO (with optimal regularization parameter). The problem configuration, i.e., $(\delta,\rho)$ lies above $\ell_1$ phase transition. {\em DeepCodec} significantly outperforms LASSO in this setting.}
\label{fig:sample}
\end{center}
\end{figure}

Figure \ref{fig:sample} shows examples of signal recoveries using {\em DeepCodec} and LASSO. The original signal is a $k$-sparse signal in $\bR^N$ where $k=64$ and $N=512$. {\em DeepCodec} recovers this signal from $M=64$ measurements. i.e., $(\delta,\rho)=(0.125,1)$ while LASSO recovers it from $M=154$ measurements, i.e., $(\delta,\rho)=(0.3,0.42)$. Therefore, as we can see in this Figure, {\em DeepCodec} has solved a more challenging recovery problem significantly better than the LASSO with optimal regularization parameter. 

\section{Conclusions}\label{sec:con}
In this paper we have developed {\em DeepCodec} that is a novel computational sensing framework for sensing and recovering structured signals. We have shown that {\em DeepCodec} can learn to take undersampled measurements and recover signals from undersampled measurements using convolutional and sub-pixel convolution layers. We compared  {\em DeepCodec} with $\ell_1$-minimization from the phase transition point of view and showed that it significantly outperforms $\ell_1$-minimization in the failure side of phase transition. In addition, compared to using random measurements we experimentally showed how learning undersampled measurements enhances the overall recovery performance, speeds up training of recovery framework, and could lead to having fewer parameters to learn.

\begin{table}[]
\centering
\caption{Average NMSE of test set signals for all three methods. {\em DeepCodec} outperforms {\em DeepInverse} and the LASSO in all cases in spite of the fact that it is solving a significantly more difficult recovery problem in each case.}
\label{tab:nmse}
\begin{tabular}{| c | c | c | c | c | c |}
\hline
                & \multicolumn{1}{l|}{LASSO} & \multicolumn{2}{c||}{DI} &                 & \multicolumn{1}{c|}{DeepCodec} \\ \hline \hline
$(\delta,\rho)$ & NMSE                       & \multicolumn{2}{c||}{NMSE}        & $(\delta,\rho)$ & NMSE                                \\ \hline
(0.3,0.42)      & 0.0466                     & \multicolumn{2}{c||}{0.0140}      & (0.125,1)       & \textbf{0.0136}                     \\ \hline
(0.5,0.56)      & 0.0312                     & \multicolumn{2}{c||}{0.0112}      & (0.25,1.12)     & \textbf{0.0110}                     \\ \hline
(0.7,0.72)      & 0.0164                     & \multicolumn{2}{c||}{0.0104}      & (0.5,1.003)     & \textbf{0.0052}                     \\ \hline
\end{tabular}
\end{table}

\begin{table}[]
\centering
\caption{Number of parameters that {\em DeepCodec} and {\em DeepInverse} learn in each problem instance. {\em DeepCodec} uses significantly fewer number of parameters while outperforming {\em DeepInverse} in all the cases.}
\label{tab:numParams}
\begin{tabular}{|c|c||c|c|}
\hline
\multicolumn{2}{|c||}{DeepInverse}         & \multicolumn{2}{c|}{DeepCodec}                            \\ \hline \hline
$(\delta,\rho)$ & $\#$parameters & $(\delta,\rho)$ & \multicolumn{1}{l|}{$\#$parameters} \\ \hline
(0.3,0.42)      & 198000         & (0.125,1)       & \textbf{6664}                                \\ \hline
(0.5,0.56)      & 198000         & (0.25,1.12)     & \textbf{5096}                                \\ \hline
(0.7,0.72)      & 198000         & (0.5,1.003)     & \textbf{4312}                                \\ \hline
\end{tabular}
\end{table}

\begin{table}[h!]
\centering
\caption{Computational complexity for every typical iteration/layer of different methods. Since {\em DeepCodec} recovers signals mainly in the measurements space, it is computationally cheaper than {\em DeepInverse}.}
\label{tab:runtime}
\begin{tabular}{|c|l|l|}
\hline
\multicolumn{3}{|c|}{Computational Complexity}                                                 \\ \hline \hline
LASSO                                   & \multicolumn{1}{|c|}{DeepInverse} & \multicolumn{1}{c|}{DeepCodec} \\ \hline
\multicolumn{1}{|l|}{$\mathcal{O}(MN)$} & $\mathcal{O}(N)$        & $\mathcal{O}(M)$           \\ \hline
\end{tabular}
\end{table}

\balance
\bibliographystyle{IEEEbib}
\bibliography{root}

\end{document}